# A Data Balancing and Ensemble Learning Approach for Credit Card Fraud Detection

Yuhan Wang
Columbia University
New York, USA

*Abstract-* **This research introduces an innovative method for identifying credit card fraud by combining the SMOTE-KMEANS technique with an ensemble machine learning model. The proposed model was benchmarked against traditional models such as logistic regression, decision trees, random forests, and support vector machines. Performance was evaluated using metrics, including accuracy, recall, and area under the curve (AUC). The results demonstrated that the proposed model achieved superior performance, with an AUC of 0.96 when combined with the SMOTE-KMEANS algorithm. This indicates a significant improvement in detecting fraudulent transactions while maintaining high precision and recall. The study also explores the application of different oversampling techniques to enhance the performance of various classifiers. The findings suggest that the proposed method is robust and effective for classification tasks on balanced datasets. Future research directions include further optimization of the SMOTE-KMEANS approach and its integration into existing fraud detection systems to enhance financial security and consumer protection.**

## I. Introduction

As the Internet and information technology continue to evolve, e-commerce platforms have gained widespread adoption, leading to a rapid increase in the use of credit card payments for online shopping. This has resulted in an exponential increase in the number of fraudulent credit card transactions and significant financial losses for consumers and financial institutions.

Credit card fraud refers to the act of obtaining improper benefits without the authorization of the cardholder. Common types of credit card fraud include stolen card transactions and credit card cashing [1]. With the rapid development of online transactions, an effective fraud detection method is necessary [2]. Traditional data statistical analysis methods, which compare users' consumption records to assess their credit scores, have several disadvantages, such as high false alarm rates and an inability to capture dynamic changes in user behavior.

For the serious imbalance of credit card fraud data, existing solutions mainly use data enhancement, unbalanced learning, and integrated learning methods. Traditional methods employ rudimentary resampling techniques that solely depend on sample resampling or the absence of representative synthetic samples. Consequently, these methods introduce bias in the classifier's posterior probability during training.

This paper aims to improve two main issues in current research. On one hand, it can detect balance and missing values of credit card datasets, inspect the features of credit card datasets, and resample the datasets. On the other hand, it processes the samples in a balanced manner through the SMOTE-KMEANS algorithm [3]. In conclusion, the refined dataset, comprising the balanced samples and selected features, is utilized to train an ensemble-based predictive model on the test set. This approach results in a notable boost in the precision and overall efficacy of detecting credit card fraud.

## II. Background

The detection of credit card fraud has been extensively studied using machine learning and deep learning techniques. Traditional fraud detection approaches often suffer from high false positive rates due to the imbalanced nature of datasets. Recent advancements in ensemble learning, feature engineering, and oversampling techniques have significantly improved the performance of fraud detection models.

Several studies have explored the role of neural networks in financial anomaly detection. Esenoghlo et al. [4] proposed a neural network ensemble model with feature engineering to enhance credit card fraud detection, demonstrating improvements in classification accuracy. Similarly, deep learning techniques such as convolutional neural networks (CNNs) have been applied to financial forecasting, where Liu [5] developed improved CNN architectures to predict stock market volatility. In the context of financial statement anomaly detection, Du [6] introduced an optimized CNN model, showing the potential of deep learning in financial fraud identification.

To address the challenge of imbalanced datasets, various resampling and data augmentation techniques have been proposed. Levy et al. [7] examined threshold optimization and random undersampling methods to mitigate class imbalance in credit card fraud detection, leading to enhanced model performance. Wang [8] introduced a Markov network-based classification method with adaptive weighting to further tackle the imbalance issue in financial datasets. Moreover, Qi et al. [9]

leveraged graph neural networks for hierarchical mining of complex imbalanced data, presenting a structured approach for data balancing.

Beyond supervised learning approaches, reinforcement learning has been explored in financial fraud detection. Yao [10] introduced a time-series nested reinforcement learning framework for dynamic risk control in financial markets, demonstrating its capability to adaptively respond to market fluctuations. Jiang et al. [11] applied Q-learning for asset allocation and dynamic risk control, highlighting the effectiveness of reinforcement learning in financial decision-making. Temporal and probabilistic modeling techniques have also been utilized to improve fraud detection performance. Zhou et al. [12] investigated the application of temporal convolutional networks (TCNs) for high-frequency trading (HFT) in blockchain markets, offering insights into market signal prediction. Du et al. [13] proposed a structured reasoning framework using probabilistic models for unbalanced data classification, which aligns with the objective of detecting fraudulent transactions in financial systems.

Contrastive learning and feature fusion techniques have been introduced to enhance data representations in imbalanced learning problems. Hu et al. [14] applied contrastive learning for cold start recommendation, demonstrating how adaptive feature fusion can improve classification tasks with limited labeled data. Additionally, Feng et al. [15] integrated deep learning with ResNeXt-based collaborative optimization for financial data mining, suggesting potential improvements in credit card fraud detection through advanced feature extraction.

Ensemble learning has been widely utilized to improve fraud detection performance. Long et al. [16] designed an adaptive transaction sequence neural network to enhance money laundering detection, showcasing the effectiveness of deep learning-based ensembles in financial security applications. Similarly, Swart et al. [17] combined neural network ensembles with feature engineering, reinforcing the advantages of ensemble techniques in handling fraudulent transactions. Audit fraud detection has also been investigated using advanced deep learning models. Du [18] introduced an efficiency-driven neural network that integrates separable convolution and self-attention mechanisms for enhanced fraud detection in auditing. This approach underscores the growing importance of attention-based architectures in financial anomaly detection.

The current algorithm processing framework for credit card fraud detection mainly includes the preprocessing stage, feature selection stage, data resampling stage, and ensemble model construction stage. The specific credit card fraud detection process is shown in Figure 1.

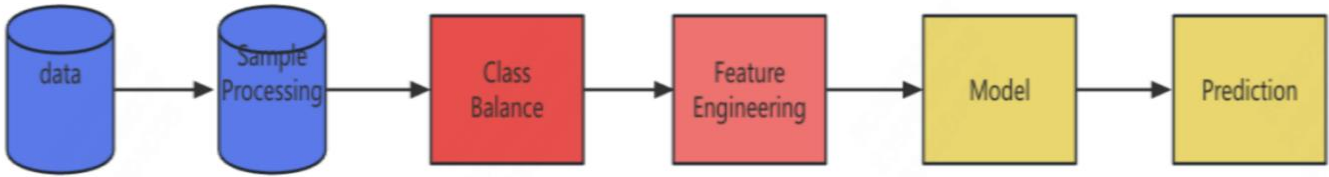

Figure 1. Credit Card Detection Framework

## III. METHOD

The proposed credit card fraud detection approach uses a SMOTE-KMEANS deep learning ensemble, integrating SMOTE, KMEANS, BiLSTM, CNN, and XGB to construct a Bagging-based ensemble framework. Figure 2 shows the flowchart of the proposed method. The process starts by applying the improved SMOTE- KMEANS algorithm to the severely imbalanced original dataset to achieve a balanced dataset [19]. Fraudulent transactions are then detected through a bagging deep learning ensemble framework, using Bi-LSTM, Bi-GRU, and CNN as base classifiers for out-of-sample predictions. These predictions, paired with actual labels, create a new dataset for training an XGB-based meta-classifier.

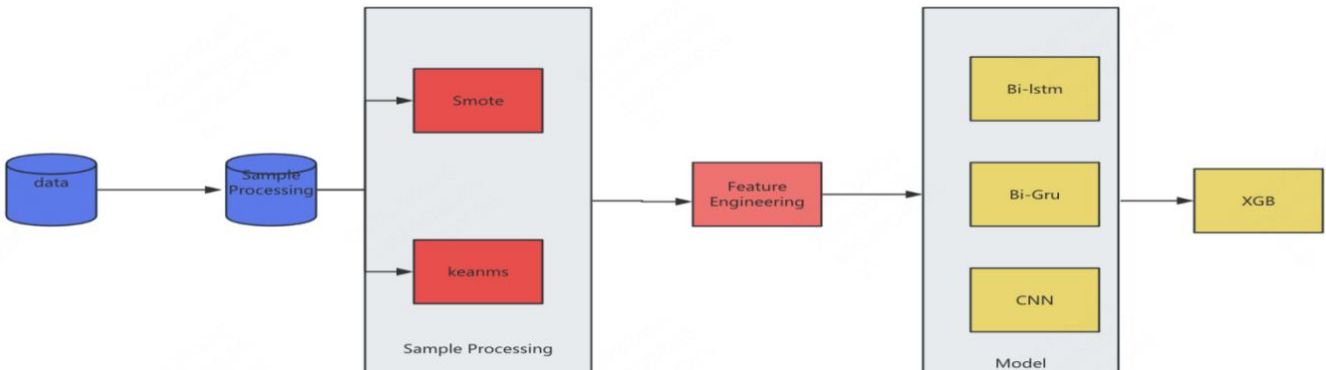

Figure 2. Ensemble Learning Framework Based on Bagging Deep Learning

## A. SMOTE-KMEANS Algorithm

To address the unbalanced category problem, training data is required for the ensemble framework. Fraudulent transactions often make up less than 1% of the data. This number of fraudulent instances in the training set is crucial to tackle category imbalance. SMOTE is a widely used oversampling method that resolves imbalanced classification by generating new samples for minority classes from each minority sample and its nearest neighbors [20]. Each synthetic sample is calculated as:

$$x_s = x_i + rand(0,1) * |x_i - x_l| \quad (3\text{-}1)$$

In Equation (3-1), $x_i$ is a given sample of a few classes, $x_l$ represents the random sample selected from the K nearest neighbors of $x_i$, and rank (0,1) is the random number between 0 and 1. To enhance SMOTE, this paper combines it with the Kmeans algorithm, presenting an improved SMOTE algorithm (Figure. 3).

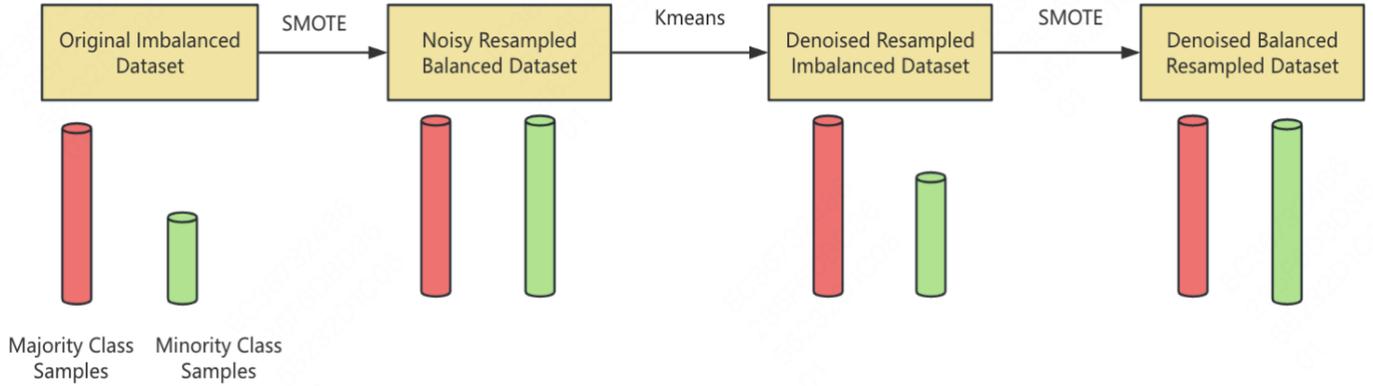

Figure 3. Smote-Kmeans algorithm

SMOTE oversampling avoids overfitting by creating new minority class samples. However, SMOTE inherent stochasticity introduces noisy samples, leading to more noise propagation. Additionally, SMOTE ignores the majority class's nearest neighbors, causing class overlap between generated minority samples and majority samples. Using SMOTE with noise removal techniques reduces noise from SMOTE or the original dataset, enhancing classifier performance.

The dataset was divided using 10-fold cross-validation. One fold was reserved for model evaluation, and the remaining nine folds were used for training. SMOTE was first applied to the training set to create an oversampled and potentially noisy dataset. Then, Kmeans clustering removed detected noise, resulting in a clean but still imbalanced dataset merged with the original training data. Finally, SMOTE was applied again to achieve a clean, balanced dataset.

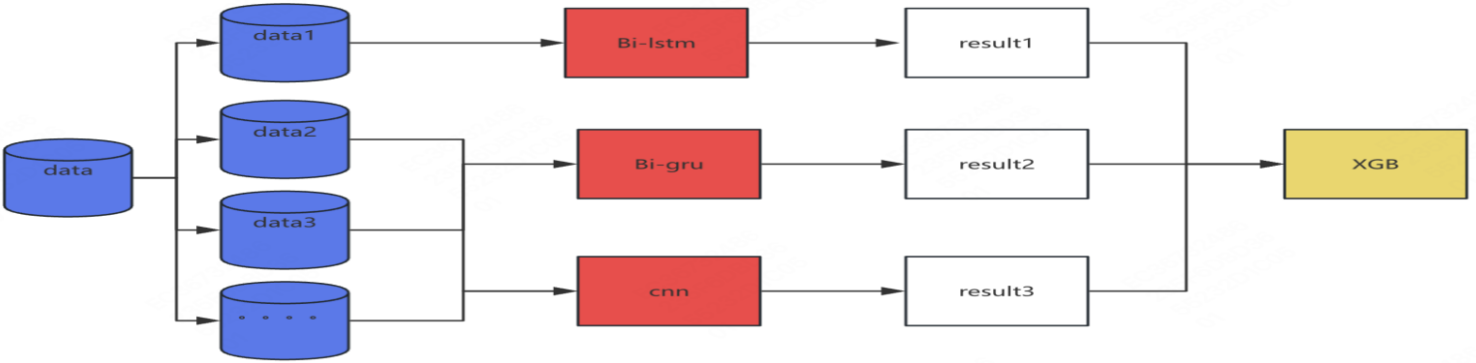

Figure 4. Ensemble Learning Based on Bagging

The k-means clustering algorithm minimizes the within-cluster sum of squares (WCSS), defined as:

$$J = \sum_{i=1}^{k} \sum_{x \in s_i} ||x - u_i||^2 \quad (3\text{-}2)$$

where $s_i$ is cluster $i$, $u_i$ is the centroid of cluster $i$, and $k$ is the number of clusters. The k-means goal is to find a partitioning that minimizes *J*, resulting in compact and separate clusters. This method is efficient for large datasets but requires a predefined k and may be sensitive to initial conditions and outliers.

## B. Ensemble Learning Based on Bagging

This paper presents an ensemble learning method based on bagging, which is a machine learning approach that integrates multiple algorithms to achieve superior classification performance compared to single-base models. Individual machine learning classifiers often face challenges like high bias, classifiers to enhance detection performance. In this study, we construct an integrated model framework by combining Bi-LSTM, Bi-GRU, CNN, and XGBoost. The specific process is shown in Figure. 4.

LSTM is an advanced recurrent neural network architecture widely used to avoid long-term dependencies and overcome gradient problems in traditional RNNs, improving long-term dependency handling [21].

The Gated Recurrent Unit (GRU) is a type of recurrent neural network specifically designed to address potential issues such as gradient problems. It features a reset gate and an update gate, and it combines the cell state C with the hidden state h. The respective formulas are as follows:

$$z_t = \sigma(W_z[h_{t-1}, x_t] + b_z) \quad (3\text{-}3)$$

$$r_t = \sigma(W_r[h_{t-1}, x_t] + b_r) \quad (3\text{-}4)$$

$$\tilde{h}_t = \tanh(W_c[r_t \odot h_{t-1}, x_t] + b_c) \quad (3\text{-}5)$$

$$h_t = (1 - z_t) \odot h_{t-1} + z_t \odot \tilde{h}_t \quad (3\text{-}6)$$

In Equation (3-5), $h_t$ represents the current state, $b_c$ represents the bias term, $h_{t-1}$ represents the state from the previous time step, $w_c$ represents the weight matrix, $x_t$ represents the input at this time step.

CNN are deep learning architectures that excel at handling data with grid-like structures, such as images. They utilize convolutional layers to extract hierarchical features from the input, employing filters that move across the input data to identify patterns such as edges and textures. CNNs also include pooling layers to decrease dimensionality and fully connected layers for performing classification tasks. Their capability to learn spatial hierarchies makes them particularly effective for image recognition and computer vision tasks.

XGBoost, or eXtreme Gradient Boosting, is a powerful machine learning algorithm frequently used in both classification and regression problems [22]. It employs an ensemble technique with decision trees, where each new tree is designed to address the errors made by its predecessors, resulting in a highly accurate predictive model. Known for its robust mathematical foundation and innovative advancements, XGBoost excels at processing large datasets swiftly and accurately. It incorporates regularization methods such as L1 and L2 to reduce overfitting and improve model generalization. Furthermore, XGBoost offers parallel processing capabilities, allowing it to utilize multiple CPU cores to accelerate training on extensive datasets.

## IV. EXPERIMENT

### A. Dataset

For the experimental dataset, the European Credit Card Fraud Transaction public dataset [23] is utilized, which can be directly downloaded from Kaggle. This dataset comprises data from 284,807 credit card transactions, including 492 cases of fraud. It exhibits a significant class imbalance, with fraudulent transactions representing only 0.172% of the total transactions. The dataset consists of 30 features: transaction time and amount, along with an additional 28 features derived from PCA for dimensionality reduction and anonymization to protect privacy. The time feature indicates the interval between the current transaction and the initial transaction. The amount features include the transaction amount and the class label, where a class label of 1 signifies a fraudulent transaction, and 0 signifies a legitimate transaction.

### B. Experiment Results Analysis

The proposed model was compared with several existing models, including logistic regression, decision tree (DT), random forest (RF), and support vector machine (SVM), all of which were implemented using the scikit-learn library. The evaluation metrics used for comparison were recall, accuracy, and AUC score. Table 1 provides a detailed performance analysis of each model on the original dataset.

Table 1. Metric of experiments

| Method | Accuracy | Recall | AUC |
|---|---|---|---|
| DT | 0.70 | 0.62 | 0.68 |
| RF | 0.83 | 0.89 | 0.82 |
| SVM | 0.69 | 0.70 | 0.76 |
| Ours | 0.88 | 0.90 | 0.89 |

The experimental results in Table 1 show a comparison of machine learning models based on accuracy, recall, and AUC. The DT model achieved an accuracy of 0.70, a recall of 0.62, and an AUC of 0.68. The RF model demonstrated superior performance with an accuracy of 0.83, a recall of 0.89, and an AUC of 0.82. The SVM model yielded an accuracy of 0.69, a recall of 0.70, and an AUC of 0.76. The Ensemble method outperformed all other models, achieving the highest scores across all metrics: an accuracy of 0.88, a recall of 0.90, and an AUC of 0.89. The Ensemble method stands out as the most effective approach for this dataset, followed closely by Random Forest. These results highlight the importance of model selection and the potential benefits of using ensemble techniques to improve predictive accuracy and robustness.

Subsequently, the original SMOTE algorithm, the SMOTE-SSAE algorithm, and the proposed SMOTE-KMEANS algorithm were applied to evaluate the performance of the benchmark classifier and our method. It can be seen that all data augmentation algorithms can effectively improve the performance of different classifiers. Among them, the proposed improved SMOTE algorithm has the most obvious performance improvement for each classifier, which better maintains the diversity of categories by combining KMEANS to eliminate the noise and over-generated samples introduced by SMOTE. In addition, KMEANS can identify and handle outliers, thereby enhancing the robustness to abnormal samples in the dataset.

Table 2. AUC results for the balanced dataset

| Method | Smote | Smote-SSAE | Smote-Kmeans |
|---|---|---|---|
| DT | 0.86 | 0.88 | 0.90 |
| RF | 0.90 | 0.89 | 0.92 |
| SVM | 0.88 | 0.88 | 0.90 |
| Ours | 0.92 | 0.94 | 0.96 |

Table 2 presents a comparison of the AUC performance of various machine learning methods on an imbalanced dataset, evaluated using three oversampling techniques: SMOTE, SMOTE-SSAE, and SMOTE-KMEANS. Our proposed method consistently attains the highest AUC scores across all techniques, reaching a peak of 0.96 with SMOTE-KMEANS. This superior performance highlights the robustness of our method in classification tasks.

## V. CONCLUSION

The experimental outcomes demonstrate the superior performance of the proposed model in the domain of credit card fraud detection. The integration of the proposed model with the SMOTE-KMEANS algorithm achieved the most notable results, with an AUC score reaching 0.96. This indicates a significant advancement in accurately identifying fraudulent transactions while maintaining a high level of precision and recall. Looking ahead, future research may concentrate on further refining the SMOTE-KMEANS approach to enhance the model's predictive capabilities. This could involve fine-tuning the algorithm's parameters or incorporating more sophisticated methods to manage noise and outliers. Moreover, the potential integration of this model into existing credit card fraud detection systems could lead to a more robust and automated framework for identifying and preventing fraudulent activities. This would not only enhance the security of financial transactions but also contribute to protecting consumer interests and reducing financial losses.

Considering the critical nature of credit card fraud detection, it is imperative to address the ethical considerations associated with the deployment of automated detection systems. The model should be designed to ensure transparency in its decision-making process, allowing for human oversight to intervene when necessary. This approach will help to prevent any potential misuse of the system and uphold ethical standards while leveraging the power of AI for fraud prevention.